# Image-based Artificial Intelligence empowered surrogate model and shape morpher for real-time blank shape optimisation in the hot stamping process

Haosu Zhou, and Nan Li*

*Dyson school of Design Engineering, Imperial College London*

n.li09@imperial.ac.uk

Abstract:

As the complexity of modern manufacturing technologies increases, traditional trial-and-error design, which requires iterative and expensive simulations, becomes unreliable and time-consuming. This difficulty is especially significant for the design of hot-stamped safety-critical components, such as ultra-high-strength-steel (UHSS) B-pillars. To reduce design costs and ensure manufacturability, scalar-based Artificial-Intelligence-empowered surrogate modelling (SAISM) has been investigated and implemented, which can allow real-time manufacturability-constrained structural design optimisation. However, SAISM suffers from low accuracy and generalisability, and usually requires a high volume of training samples. To solve this problem, an image-based Artificial-intelligence-empowered surrogate modelling (IAISM) approach is developed in this research, in combination with an auto-decoder-based blank shape generator. The IAISM, which is based on a Mask-Res-SE-U-Net architecture, is trained to predict the full thinning field of the as-formed component given an arbitrary blank shape. Excellent prediction performance of IAISM is achieved with only 256 training samples, which indicates the small-data learning nature of engineering AI tasks using structured data representations. The trained auto-decoder, trained Mask-Res-SE-U-Net, and Adam optimiser are integrated to conduct blank optimisation by modifying the latent vector. The optimiser can rapidly find blank shapes that satisfy manufacturability criteria.  As a high-accuracy and generalisable surrogate modelling and optimisation tool, the proposed pipeline is promising to be integrated into a full-chain digital twin to conduct real-time, multi-objective design optimisation.

# 1. Introduction

Though numerical methods, such as Finite Element Analysis [1], have accelerated design validation, engineers still rely on empirical and trial-and-error iterations to conduct design optimisation. This problem becomes more serious in design-for-manufacturing, since design engineers usually lack expertise in manufacturing processes. For example, in the design of a hot-stamped B-pillar, a variety of manufacturing factors need to be carefully tuned to ensure manufacturability [2][3][4]. A minor change of the shape or processing parameters might lead to significant change of the manufacturability metrics, such as maximum thinning/thickening [5]. As a result, the traditional trial-and-error design optimisation process is time-consuming and unreliable.

For more efficient and reliable design optimisation, engineers expect real-time feedback of manufacturability and intelligent design iterations. To achieve this goal, surrogate modelling provides an effective enabler, such as response surface method [6], radial basis function [7], Kriging (also known as Gaussian process) [8][9][10], and neural network [11], etc. A surrogate model is trained using a number of simulation samples. Based on the trained surrogate model as a real-time simulator, the optimisation algorithm can efficiently search for more performant design candidates, without requiring expensive simulation.

Given the benefits mentioned above, the traditionally scalar-based Artificial Intelligence empowered surrogate model (SAISM) has been commonly used [12-14]. Though SAISM has been widely applied to design optimisation tasks, its performance can only be guaranteed when there is enough data available and the mapping relationship to be learned is relatively simple. When the design optimisation task is complicated, especially when under multiple parameterisations, SAISM suffers from shortcomings of low accuracy, poor generalizability, lack of informativeness, and high-volume data requirement. The reasons for these shortcomings have been attributed to the information loss caused by the scalar-based data representations of SAISM, at both input and output sides [15].

To overcome the shortcomings of SAISM, the image-based Artificial Intelligence empowered surrogate model (IAISM) has been increasingly investigated, which leverages more information contained in the shapes as inputs and physical fields as outputs [16-18]. IAISM overcomes the shortcomings of SAISM using a structured and informative-preserving data representation, thus can lead to better accuracy and generalizability. The development and implementation of IAISM emerged in the field of fluid dynamics, mainly because Eulerian meshing in fluid dynamics can naturally be pixelized [19-23]. Based on Eulerian configurations, IAISM has also been extended to solid mechanics [24-32], additive manufacturing [33-37], and topology optimisation [38-43]. Recently, IAISM have been specialized to sheet stamping and bending [44-53]. This is because the inputs and outputs in these scenarios can be represented by 2D images or fields. The advantages of IAISM over SAISM and corresponding reasons have been demonstrated by Zhou et al., especially the small data requirement of IAISM [15][54]. To extend IAISM to general design optimisation cases, especially for arbitrary 3D shapes, implicit shape representation and graph neural network are being investigated [11][55-58]. By representing complex shapes in a uniform latent space and mapping the physical fields onto regularly aligned planes, arbitrary shapes and physical fields can be uniformly processed using a neural network with a fixed architecture. Despite of these advances, few studies have conducted a systematic comparison between SAISM and IAISM based on a benchmark dataset. As IAISM does not directly predict the scalar indicators, such as maximum thinning/thickening, it is still unclear whether and how much IAISM can be significantly superior to SAISM, especially under multiple parameterisations and small data.

To fill this gap, in this article, a study to allow comprehensive comparison between SAISM and IAISM will be conducted based on a blank optimisation case study with multiple blank shape parameterisations and a limited amount of simulation data. In section 2, the blank shape optimisation problem, the workflows of SAISM and IAISM, as well as the techniques used for SAISM (including

Kriging and response surface function) and IAISM (a special architecture of convolutional neural network, named as Mask-Res-SE-U-Net) will be introduced. In section 3, the latent representation of blank shapes and the data sampling strategy, along with training and evaluation of both SAISM and IAISM, will be explained/demonstrated. In section 4, the results of the SAISM and IAISM will be compared and discussed. In section 5, conclusions will be stated, and the outlook of further development of IAISM will be provided.

## 2. Workflows and techniques

In this section, the workflows and techniques of SAISM/IAISM will be introduced based on a blank shape optimisation problem, which aims to improve the manufacturability of a B-pillar component. In section 2.1, the blank optimisation problem will be elaborated, including the PAM-STAMP simulation settings and materials properties. In section 2.2, the workflows of SAISM and IAISM will be demonstrated. The potential advantages of IAISM will be theoretically interpreted. In section 2.3, the AI techniques constituting SAISM and IAISM will be introduced, including radial basis functions, Kriging models, and neural networks.

### *2.1 Problem statement*

During the blank optimisation, hot stamping simulation is conducted in professional software programmes, such as PAM-STAMP in our research. A typical PAM-STAMP setup of a hot stamping simulation is demonstrated in Figure 1 (a). As shown in the left, a hot stamping simulation comprises five parts, the blank, die, punch, blank holder and spacer. During the hot stamping process, the punch pushes the undeformed blank into the die, and the blank holder prevents the blank edge from being overly warped. To avoid rapid quenching at the blank edge, a spacer with a small thickness is placed between the blank holder and the die. As shown in the right, two blank shapes and corresponding thinning fields are compared. In the top case, blank shape 1 is a software-unfolded blank, which does not consider manufacturing factors. The corresponding thinning field 1 indicates that blank shape 1 is non-manufacturable, as both thinning/thickening exceed the industrial criteria (0.15/0.1). In the bottom case, blank shape 2 is an optimised blank with consideration for manufacturability, and the corresponding thinning field 2 satisfies the industrial criteria (0.15/0.1). This comparison indicates the significant impact of the blank shape on manufacturability.

    In our research, the spacer thickness is set to be 3 mm. The aggregated blank holder force is set to be 200 kN. The initial blank thickness is set to be 1.6 mm. The initial blank temperature is set to be 765 ℃. The initial tool temperatures of die, punch, blank holder, and spacer are set to be 25℃. The stamping speed is fixed to be 150 mm/s. The friction coefficients of the contact pairs, which includes die/blank, punch/blank, blank holder/blank, spacer/blank, are fixed to be 0.4. To simulate the non-isothermal and visco-plastic material behaviour, the flow stress of UHSS under different temperatures and strain rates are displayed in Figure 1 (b). These flow stress curves, without considering damage models, are plotted by using a fitted unified material model developed by Lin et al [59] in MATLAB, and then imported into PAM-STAMP as lookup tables.

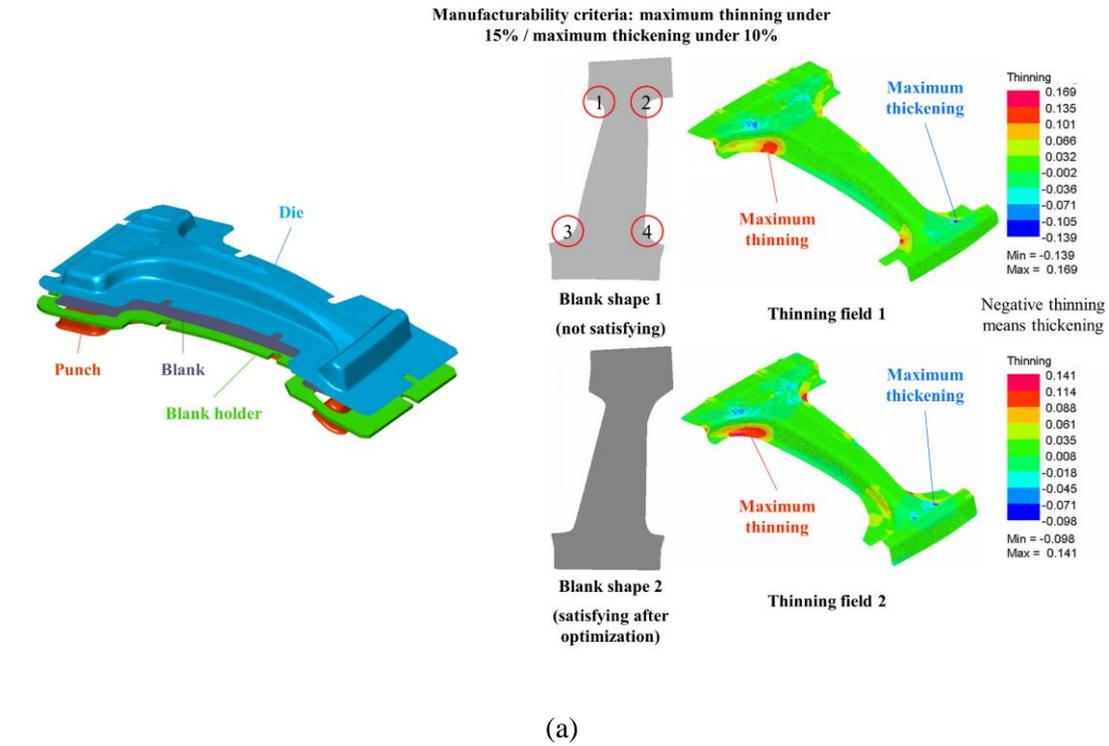

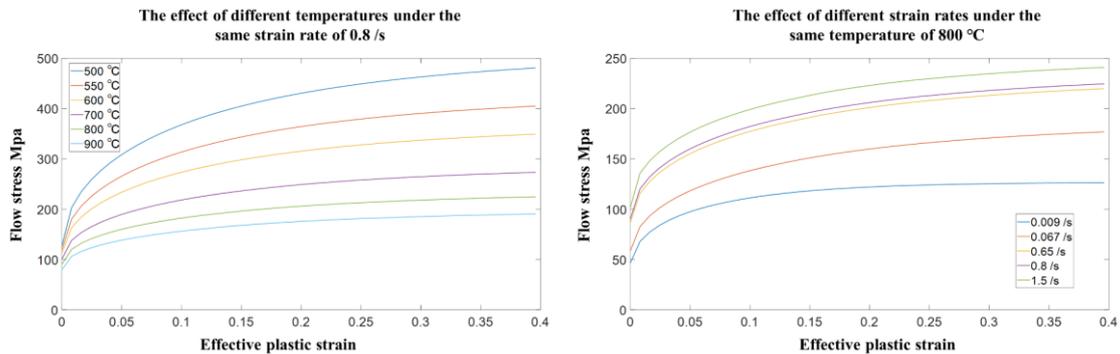

Figure 1. (a) the effect of blank shape on the as-formed thinning field; (b) the flow stress curves of ultra-high-strength steel (UHSS) under different temperatures and strain rates.

Based on the simulation tool, engineers can conduct trial-and-error blank optimisation, as shown in Figure 2 (a). At the beginning of a trial-and-error blank optimisation process, a software-unfolded blank shape is generated and parameterised. Then, the blank shape is updated and evaluated using PAM-STAMP hot stamping simulation. If the manufacturability indicators, in this case the maximum thinning/thickening, satisfy the industrial criteria (0.15/0.1 in this case), the manufacturability process will be completed, and the next design or production stage will be launched. If not, a next-round blank shape design iteration is required to modify the blank shape parameters. It should be noted that one blank shape can usually be parameterised in several ways, which will be further explained in later sections. Multi-parameterisation brings significantly more difficulties into the trial-and-error optimisation. According to industrial practice, this trial-and-error design process typically costs two weeks' work of an experienced engineer and cannot guarantee a feasible blank shape. If no manufacturable blank shape can be found, extra multi-team engineering work is required, such as tweaking other manufacturing parameters or even returning to the previous design stage to modify the B-pillar shape.

To avoid the difficulties of trial-and-error blank optimisation, surrogate-based optimisation can be applied, the general pipeline of which is demonstrated in Figure 2 (b). A well-trained and differentiable surrogate model can take arbitrary blank shape parameters as inputs and predict the corresponding manufacturability indicators in real-time, which are maximum thinning/thickening values in our research. Gradient-based optimisation can be conducted to automatically optimise the blank shape parameters until a convergence to a manufacturable blank shape. To be noted, the specific forms of blank shape parameters and manufacturability indicators depend on data representations, which will be demonstrated in section 2.2.

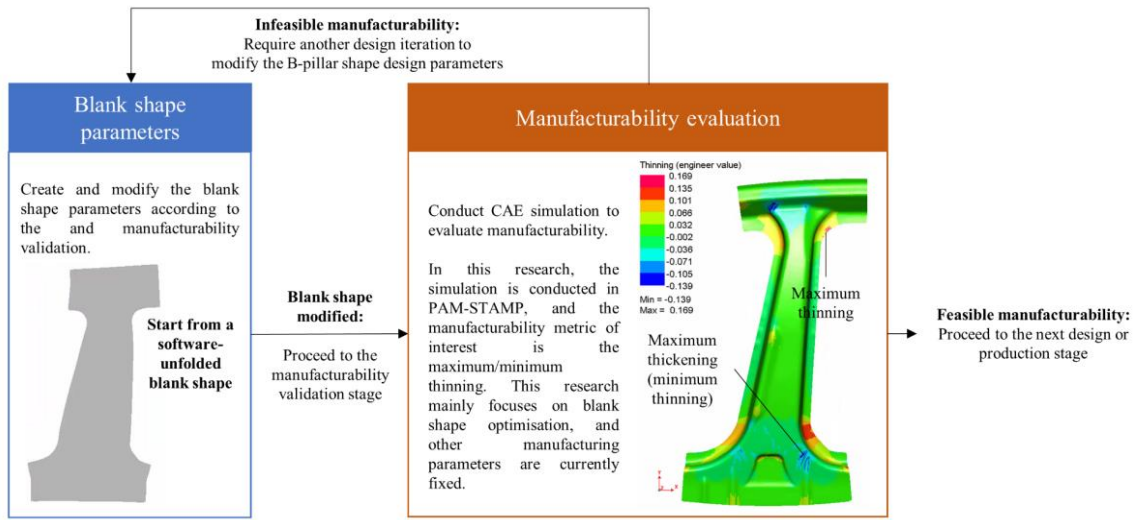

(a)

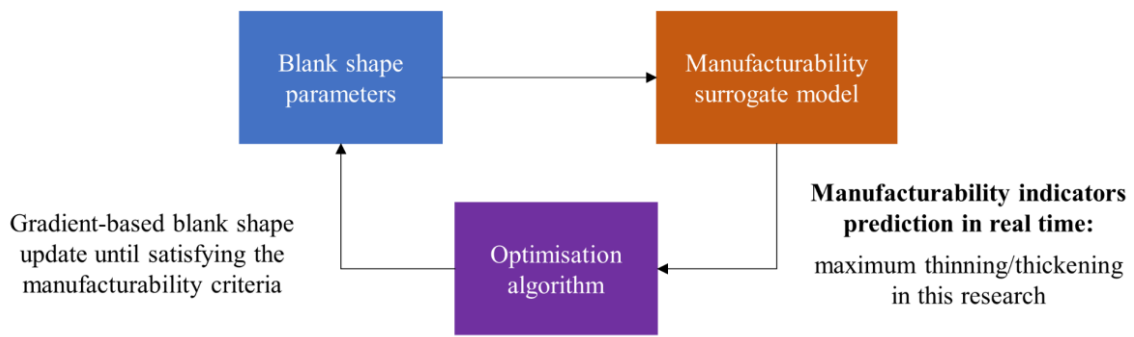

(b)

Figure 2. Manual and surrogate-based pipelines of manufacturability design for the initial blank shape modification of a B-pillar component: (a) manual trial-and-error design iterations to fulfil the manufacturability criteria; (b) surrogate-based design optimisation that can automatically search for design candidates with feasible manufacturability.

*2.2 The comparison between the workflows of SAISM and IAISM*

The reasons for these shortcomings have been attributed to the scalar-based data representations of SAISM, at both input and output sides. The blank shape optimisation case in this research is used to briefly explain these reasons. At the input side, a small local modification of the blank shape can lead to a different parameterisation, for example the region 1 in the blank shape 1 in Figure 1 (a) can be parameterised either as a rounded corner or a spline. A mapping relationship trained using scalar parameters, however, is limited to a single parameterisation. Even under a single parameterisation, scalar parameters fail to capture the intrinsic topology of a blank shape. For instance, supposing that all the four red-circled regions in the blank shape 1 in Figure 1 (a) are parameterised as rounded corners (R1, R2, R3, R4), it is obvious that R1 is spatially closer to R2 than to R3; furthermore, the topology link between R1 and R2 is different from that between R1 and R3, which can hardly be quantitatively represented using scalar parameters. At the output side, maximum thinning/thickening cannot indicate neither where these extreme material behaviours occur, nor the distributions of the thinning fields. While it is straightforward to extract the maximum thinning/thickening values from a given thinning field, it is non-trivial to determine the full thinning field from given maximum values. Considering the attributes discussed above, the shortcomings of SAISM can be qualitatively interpreted as information loss caused by scalar inputs and outputs.

The workflows of SAISM and IAISM are demonstrated in Figure 3 (a):

- For SAISM, on the input side, SAISM downscales blank shapes to scalar blank shape parameters. Explicit scalar parameters, such as the parameters used for CAD modelling, can be used when the blank shape is under a single parameterisation, while implicit scalar parameters, such as encoded latent vectors, can be used when under multiple parameterisations. On the output side, SAISM downscales full thinning fields to scalar manufacturability indicators. SAISM is trained to establish a mapping relationship between the scalar blank shape parameters and the scalar manufacturability indicators. In our research, SAISM takes latent vectors as inputs and maximum thinning/thickening values as outputs. Based on a trained SAISM, a gradient-based optimiser is applied to modify the latent vector until convergence at a satisfying design candidate.

- Instead of downscaling, IAISM, which is a convolutional neural network (CNN) [60-62], directly takes blank shapes as inputs, and full thinning fields as outputs. The maximum thinning/thickening values, which are manufacturability indicators formulating the objective function of the optimisation problem, can be extracted from the predicted thinning fields. To specify, blank shapes are represented as planar signed-distance fields (SDFs), which are demonstrated in Figure 3 (b). For a given blank shape mapped on a grid of pixels, the absolute SDF value of each pixel is confirmed by calculating its distance to the blank contour. The SDF value is positive is the pixel is in the exterior of the blank, and negative if in the interior of the blank. The gradient-based optimiser, however, cannot be directly applied to SDFs, since tuning each pixel of the SDF is time-consuming and does not guarantee reasonable optimised blank shapes. Instead, SDFs are represented using low-dimensional latent vectors, and the components of latent vectors are considered as design variables that implicitly control the blank shapes. During gradient-based optimisation, a trained decoder reconstructs the SDF from a given latent vector, and IAISM, which is trained on ground truth SDFs, predicts the full thinning field from the reconstructed SDF. The gradient-based optimiser is applied to modify the latent vector until convergence at a satisfying design candidate.

For fair comparison in our research, SAISM and FAISN share the same latent vector representation. The techniques of dimension reduction (latent representation), SAISM and IAISM will be introduced in section 2.3.

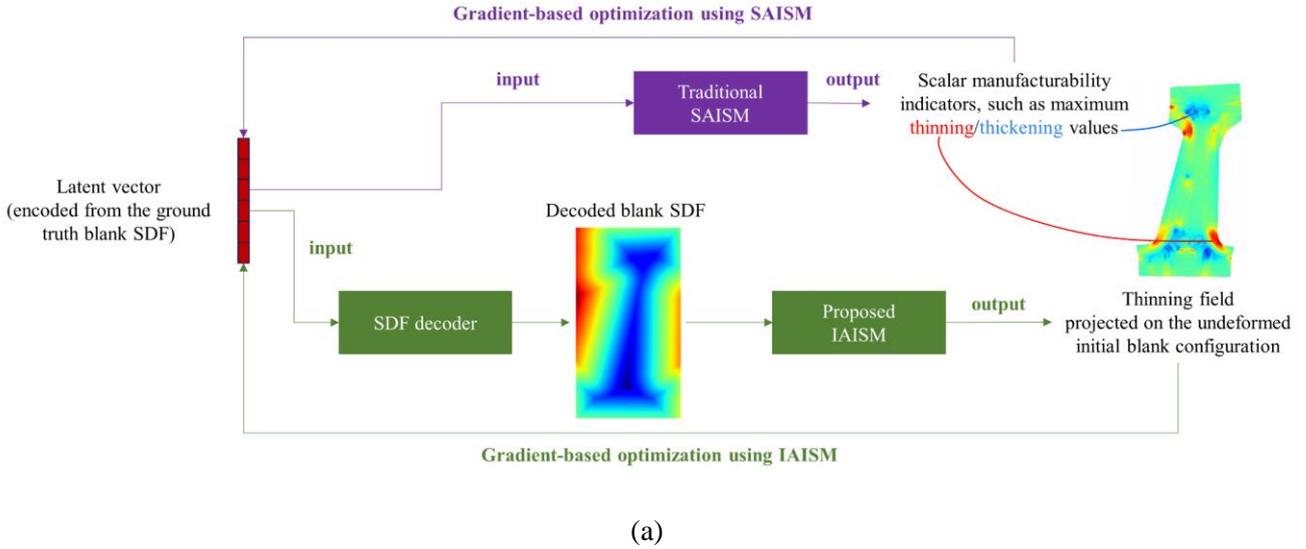

(a)

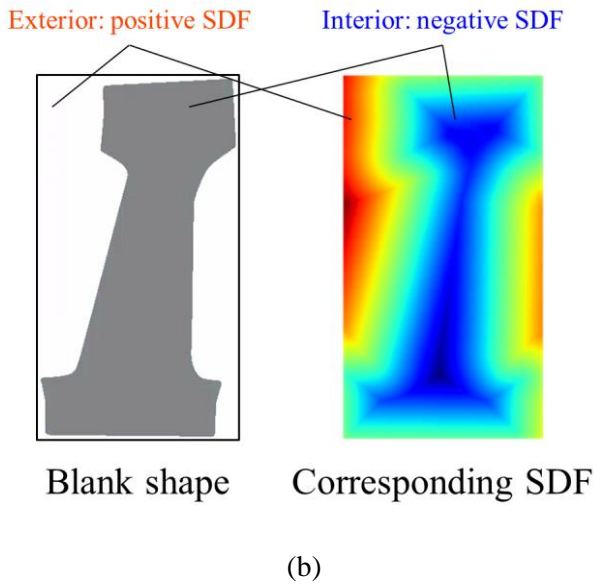

(b)

Figure 3. (a) Comparison between the workflows of traditional scalar-based Artificial Intelligence empowered surrogate models (SAISM) and proposed image-based Artificial Intelligence empowered surrogate models (IAISM); (b) the definition of signed-distance-fields of blank shapes.

*2.3 The techniques of SAISM, IAISM, and the auto-decoder*

To develop SAISM and IAISM, respectively, a variety of techniques can be implemented. For SAISM, two typical regression techniques are selected, namely the radial basis function interpolation (RBF) and Kriging interpolation. For IAISM, a convolutional neural network, with a special architecture named Mask-Res-SE-U-Net, is developed to learn a mapping relationship from SDFs to thinning fields.

In this research, both RBF and Kriging predicts the maximum thinning/thickening values of an unknown blank shape latent vector, based on a set of known latent vectors and corresponding maximum thinning/thickening values. For RBF, the known data is denoted as $\{x_j, y_j\}|_{j=1}^{N}$, and interpolates the maximum thinning or thickening values (one RBF model for one manufacturability indicator) as:

$$\hat{f}(x) = \sum_{i=1}^{N} \omega_i \varphi(\|x - x_i\|) \qquad (1)$$

$$\hat{f}(x_j) = y_j = f(x_j) \qquad (2)$$

$$\varphi(\|x - x_i\|) = \sqrt{1 + \left(\frac{\|x - x_i\|}{l}\right)^2}, \; l = \frac{1}{\binom{N}{2}} \sum_{k=1}^{N-1} \sum_{l=k+1}^{N} \|x_k - x_l\| \qquad (3)$$

In this research, multiquadric kernels are selected as shown in equation (3). When plugging equation (2) into equation (1),

$$\begin{bmatrix} \varphi_{11} & \varphi_{12} & \cdots & \varphi_{1N} \\ \varphi_{21} & \varphi_{22} & \cdots & \varphi_{2N} \\ \vdots & \vdots & & \vdots \\ \varphi_{N1} & \varphi_{N2} & \cdots & \varphi_{NN} \end{bmatrix} \begin{bmatrix} \omega_1 \\ \omega_2 \\ \vdots \\ \omega_N \end{bmatrix} = \begin{bmatrix} y_1 \\ y_2 \\ \vdots \\ y_N \end{bmatrix}, where \; \varphi_{ji} = \varphi(\|x_j - x_i\|) \qquad (4)$$

equation (4) can be denoted as $\Phi W = y$. The weights $w_i$ of each known data sample in (1) can be calculated as $W = \Phi^{-1} y$. As long as $\{x_j, y_j\}|_{j=1}^{N}$ are different from each other, $\Phi^{-1}$ is invertible. With the calculated weight matrix $W$, the maximum thinning or thickening values $\hat{f}(x)$ of an arbitrary latent vector $x$ can be predicted using equation (1). In this research, RBF is fulfilled using an open-access Python toolkit, scipy.interpolate.Rbf.

For Kriging interpolation, the known data is also denoted as $\{x_j, y_j\}|_{j=1}^{N}$, and $x_j = [x_j^1, x_j^2, \ldots, x_j^{nx}]^T$. The maximum thinning or thickening values (one Kriging model for one manufacturability indicator) of an arbitrary blank shape latent vector $x$ is interpolated as a weighted linear combination of basis functions $f_i(x)$, which is then added to a realization of a stochastic process $Z(x)$:

$$\hat{y} = \sum_{i=1}^{k} \beta_i f_i(x) + Z(x) \qquad (5)$$

$Z(x)$ is a realization of a stochastic process with mean zero and Gaussian spatial covariance function defined as:

$$cov[Z(x_i), Z(x_j)] = \sigma^2 \prod_{l=1}^{nx} \left(-\theta_l (x_l^i - x_l^j)^2\right), \forall \theta_l \in R^+ \qquad (6)$$

In the implementations, $x$ is normalized by subtracting the mean from each latent vector component and then dividing the values of each component by its standard deviation:

$$X_{norm} = \frac{X - X_{mean}}{X_{std}} \qquad (7)$$

In this research, the deterministic term $\sum_{i=1}^{k} \beta_i f_i(x)$ is replaced by a learnable constant. Kriging interpolation is fulfilled using an open-access Python toolkit, SMT.

For IAISM, a convolutional neural network (CNN) with a tailored architecture is suitable for learning an image-to-image mapping relationship. Figure 4 (a) demonstrates a Mask-Res-SE-U-Net architecture specially tailored to this study, and Table 1 specifies the network hyper-parameters. The resolution of the input blank SDF and the output predicted thinning field is $1120 \times 610$, and intermediate resolution can be calculated sequentially. The Res-SE-U-Net architecture has been developed and validated in previous studies [11][25][49][53][64-72]. Mask-Res-SE-U-Net is an enhanced version, by nullifying the out-of-blank regions using the input blank SDF. 'Res-SE' refers to a composition of residual blocks and squeeze-excitation blocks, the architectures of which are demonstrated in Figure 4 (b). In each Res-SE block, the input $x$ from the previous layer is firstly processed by two composite layers, each of which contains a convolutional layer followed by batch normalization and ReLU [73] activation. $x$ is processed to be $u_{H \times W \times C}$, where H, W, and C respectively refer to the height, weight, and channel number of the feature maps. When $u$ enters the SE block, $u$ firstly go through a global pooling layer to be $w_C$. $w_C$ is then processed by a channel-squeeze layer to be $p_{C/r}$, which contains a fully connected layer and ReLU activation. $r$ is set to be 16 in this research. $p_{C/r}$ is then processed to be $q_C$, which is the channel attention weight vector of the SE block input $u_{H \times W \times C}$, by a fully connected layer and then Sigmoid activation. $q_C$ is then replicated along the height and width directions to be $e_{H \times W \times C}$, and then multiplied by $u_{H \times W \times C}$ to be $v$. The output $v$ of the SE block is added by the input $x$ of the Res-SE block and then processed by ReLU activation. The output of this Res-SE block will be transferred to the next Res-SE block or transpose conv layer. A proper number of Res-SE blocks, which is set to be six in this research, has been approved to boost the prediction performance. To further facilitate end-to-end information flow, skip connections, which is denoted as 'U' in the overall architecture name, are implemented from the blank SDF directly to output thinning fields.

For both SAISM and IAISM, an auto-decoder [54] is developed to represent the SDFs using latent vectors. Especially for IAISM, the trained auto-decoder will be used to reconstruct the SDFs from the latent vectors during the blank optimisation. The architecture of the auto-decoder is demonstrated in Figure 4 (c), and the parameters in Table 2.

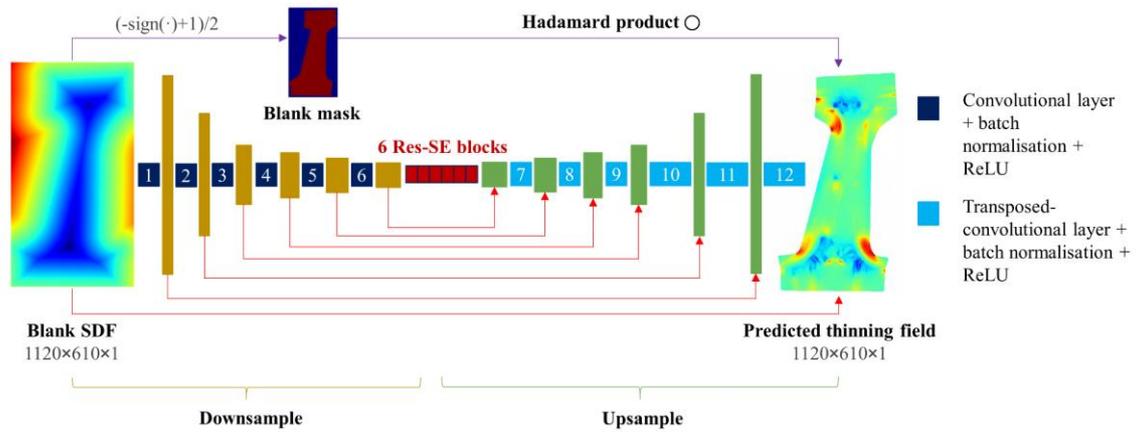

(a)

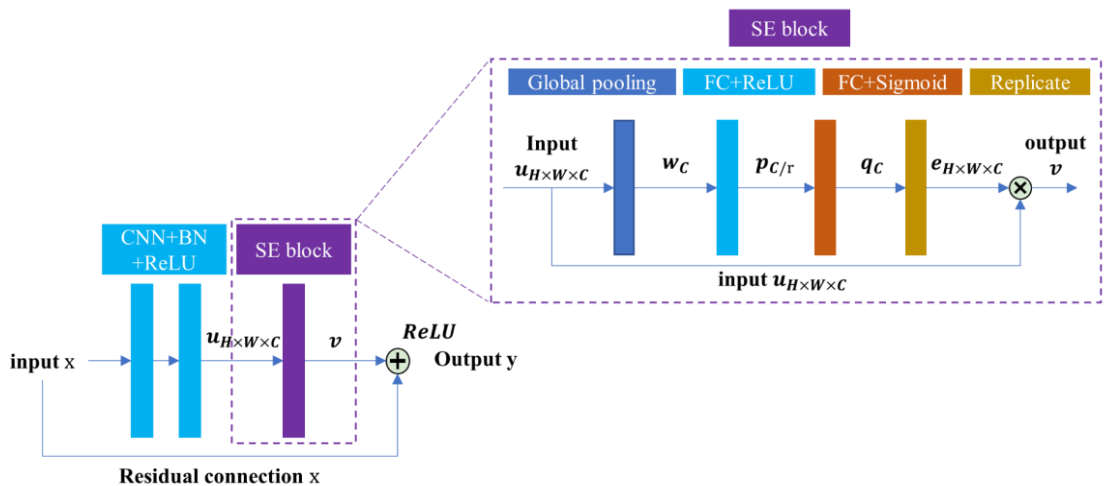

(b)

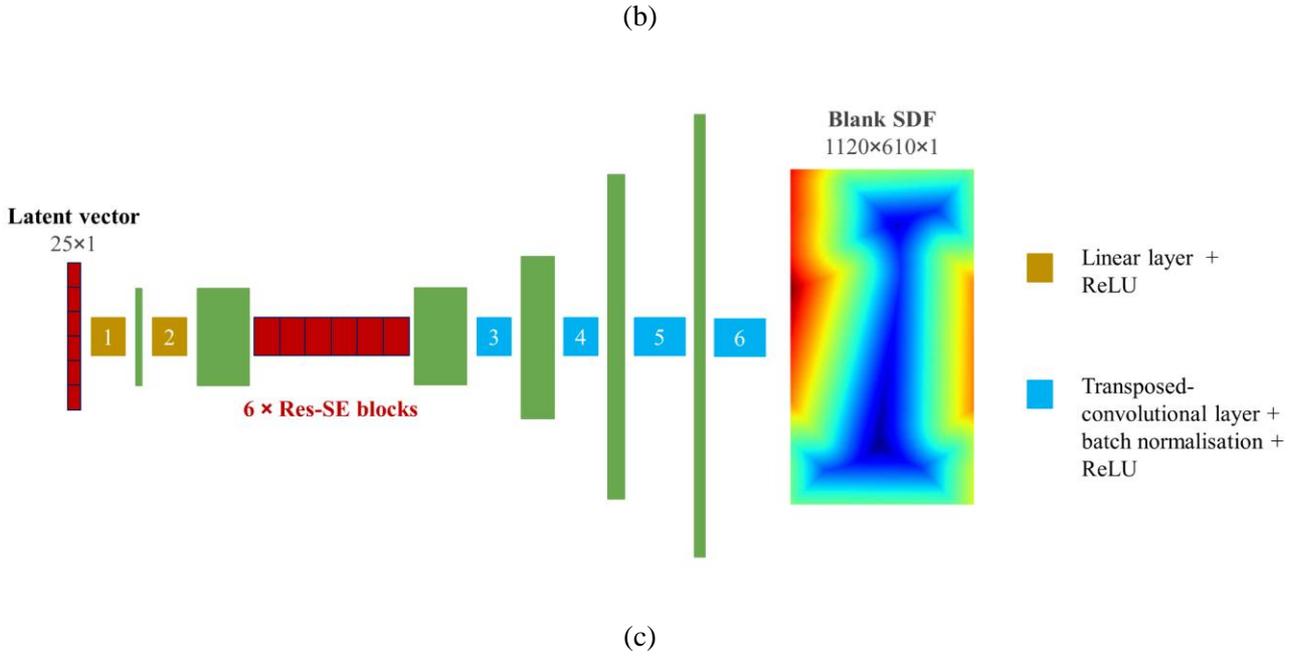

(c)

Figure 4 (a) The architecture of Mask-Res-SE-U-Net (red arrows refer to short connections and the purple arrow refers to the mask connection); (b) one Res-SE block; (c) the architecture of auto-decoder for the latent space representation and reconstruction of blank shapes.

Table 1. Mask-Res-SE-U-Net architecture parameters

| Layer | 1 | 2 | 3 | 4 | 5 | 6 | Res-SE | 7-11 | 12 |
|---|---|---|---|---|---|---|---|---|---|
| Type | Convolutional (Conv) layers | | | | | | Res-SE blocks | Transpose conv layers | Conv layers |
| Kernel | (8,8) | (8,9) | (6,5) | (4,3) | (3,3) | (3,3) | (3,3) | These layers are transposed | (5,5) |
| Stride | (2,2) | (2,1) | (2,2) | (2,2) | (2,2) | (2,2) | (1,1) | from the conv layers 2-6 with | (1,1) |
| Padding | (3,3) | (3,4) | (2,2) | (1,1) | (1,1) | (1,1) | (1,1) | the same K-S-P parameters. | (2,2) |

Table 2. Auto-decoder architecture parameters

| Layer | 1 | 2 | Res-SE | 3 | 4 | 5 | 6 |
|---|---|---|---|---|---|---|---|
| Type | Linear layer | Conv layers | Res-SE blocks | Transpose conv layers | | | |
| Kernel | | (3,3) | (3,3) | (4,3) | (6,5) | (8,9) | (8,8) |
| Stride | None | (1,1) | (1,1) | (2,2) | (2,2) | (2,1) | (2,2) |
| Padding | | (1,1) | (1,1) | (1,1) | (2,2) | (3,4) | (3,3) |

## 3. Dataset sampling and training of the auto-decoder, SAISM, and IAISM

Given the initial software-unfolded blank shape, as shown in the blank shape 1 in Figure 1 (a), surrogate optimisation is expected to rapidly search for manufacturable blank shapes by locally tailoring blank shape 1. The training and test sets should be sampled to cover the design space of local variations of blank shape 1. Based on the sampled datasets, the SAISM, IAISM, and the auto-decoder can be trained and evaluated, respectively. In this section, the blank shape parameterisation strategy, dataset generation strategy, and training parameters will be introduced.

*3.1 Blank shape parameterisation*

The software-unfolded blank shape (blank shape 1 in Figure 1 (a)) provides a starting reference point, which is reshown in Figure 5 (a). Both the maximum thinning and thickening have exceeded the industrial criteria (0.15/0.1). Considering the geometric constraints and manufacturing empirical knowledge, five sensitive regions have been circled in red across the blank shape 1, as shown in the left in Figure 5 (a). To specify, the upper edge in region 1 can be moved along its normal direction while keeping the edge as a straight line. Each of the local transition curves in regions 2 to 4 can be either parameterised as an arc or a spline. Though a spline can contain an arbitrary number of control points in theory, it is reasonable to assume that, at each region, a spline with 2 or 3 control points is sufficient to cover the possible local shape changes for this practice. An empirical parameterisation rule, which is shown in Figure 5 (b), is applied to define the shape parameters from P0 to P32. Based on the parameterisations and parameters, the local shapes from regions 1 to 5 can be completely determined. The numbers of local parameterisation in regions 1 to 5 are 1, 2, 2, 2, 2, respectively. In total, the number of overall parameterisations for the whole blank is $1\times 2\times 2\times 2\times 2 = 16$. The detailed definition of P0 to P32 is attached in Appendix A.

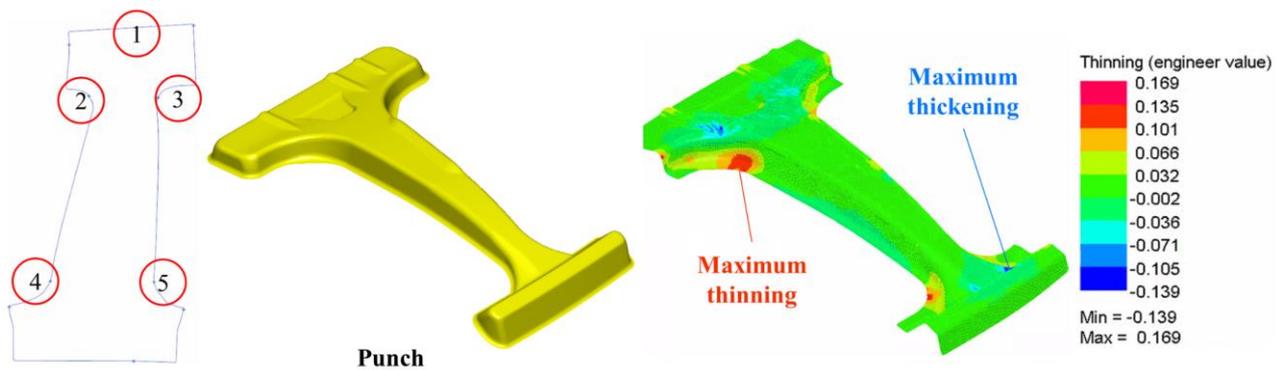

(a)

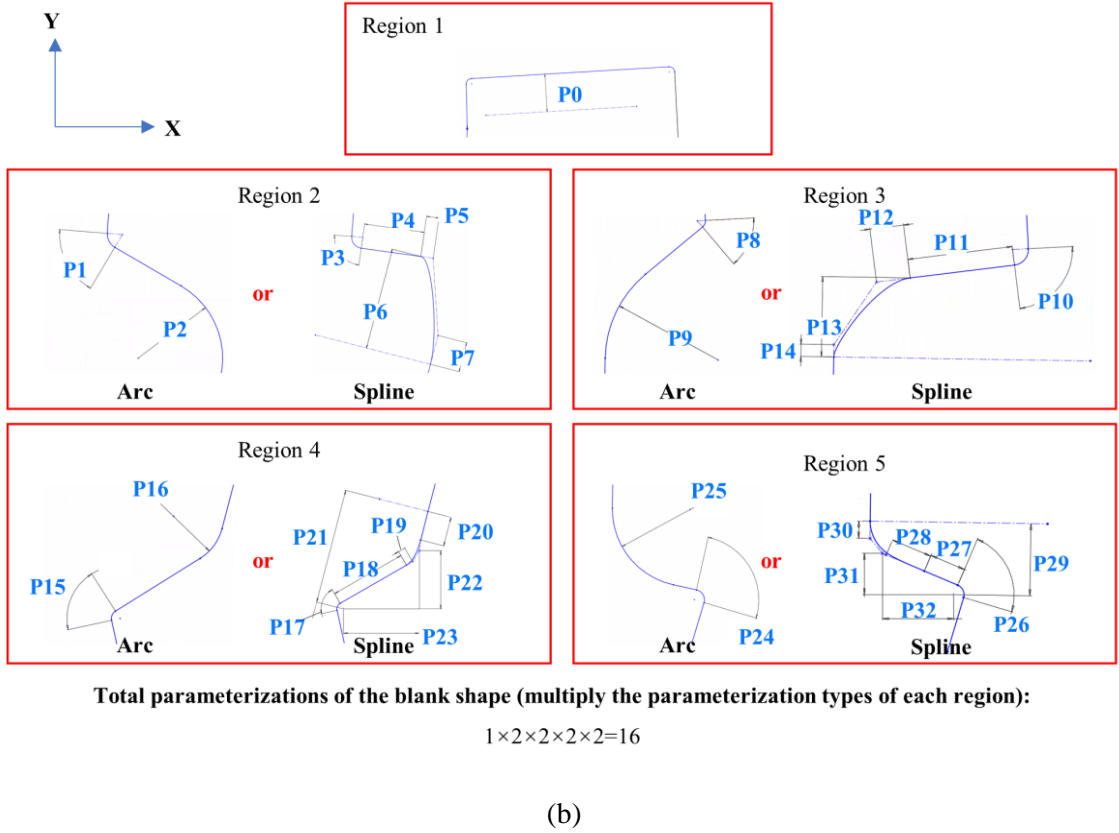

(b)

Figure 5. The PAM-STAMP simulation model and dataset sampling: (a) a software-unfolded initial blank shape without considering manufacturability and modifiable regions; (b) parameterisations for local shapes in regions 1-5, which leads to sixteen parameterisations for the blank shape.

## 3.2 Dataset generation and augmentation strategy

As the blank shape has been parameterised in Figure 5 (b), a dataset generation strategy is required to cover the aggregated design spaces of the 16 parameterisations. Though Latin-hypercube sampling (LHS) has been generally used for dataset generation, LHS cannot be directly applied in this case. This is because LHS requires the design parameters to have independent value ranges. The value ranges of design parameters from P0 to P32, however, are not independent due to the geometric complexity of the blank shape. In this case, an improved dataset generation strategy can be:

- Divide P0 to P32 into range-independent parameters (RI) and range-dependent parameters (RD). Define the ranges of the RI and define the formulas that calculate the ranges of RD with the ranges of RI as variables.
- Apply LHS to RI to generate the values of RI for each data sample.
- For each data sample, calculate the ranges of RD using the defined formulas, and apply LHS to RD.

To fulfil the improved dataset generation strategy, several empirical rules are proposed according to the geometric features of the blank shape, which is detailed in Appendix B:

Based on these empirically rules, LHS can be applied to generate the training and test sets. Instead of uniformly generating samples and then split them into training and test sets, the dataset generation strategy in this research is to uniformly generate the training set and test set, respectively This is because the former strategy might be more suitable for general AI tasks, where datasets are generated by data collection rather than data sampling. For engineering AI tasks where the design space can be explicitly defined and sampled, however, the former strategy will lead to unnecessary holes in the design space, since test samples are removed. In industrial practice, especially when under small data, the later dataset generation strategy is more suitable, which allows the training set to uniformly cover the design space.

Based on the blank parameterisation and data generation strategy, the training and test sets are generated for IAISM/SAISM and the auto-decoder, respectively:

- SAISM and IAISM share the same training and test sets, and each sample contains a blank shape and the corresponding thinning field. To verify the small-data learning nature of engineering AI in previous studies [75], a 256-sample training set and a 64-sample test set is generated for SAISM/IAISM.

- For the auto-decoder, a sample contains only a blank shape Since the auto-decoder training does not require expensive simulations, the 256 blank shapes in the IAISM/SAISM training set are added by 1024 extra blank shapes, which follow the same rules of dataset generation, to form the 1280-sample training set for the auto-decoder. The test set of the auto-decoder contains the 64 blank shapes from the test set of IAISM.

The generated training sets can be augmented to possibly improve the performance. For IAISM, data augmentation can be applied by including variations of the original training samples. As shown in Figure (6), the thinning field plotted on the initial blank, along with its corresponding blank shape, can be processed in three ways: flip horizontally, flip vertically, flip horizontally and vertically. The augmented IAISM training set contains four-fold data samples (1024) as much as the original dataset (256). For SAISM, no data augmentation is available since both inputs and outputs are scalars. For the auto-decoder, data augmentation is not applied in order to avoid generating flipped blank shapes during the optimisation.

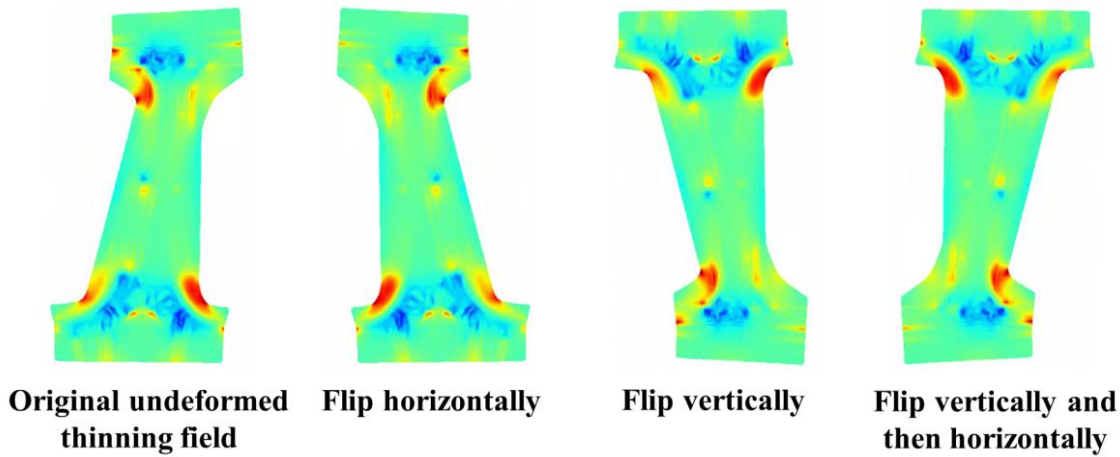

Figure 6. Data augmentation by flipping the thinning fields plotted on the initial blank in the training set.

*3.3 Training parameters*

According to the techniques introduced in section 2.3, IAISM should be directly trained on the blank shapes and physical fields; the auto-decoder should be directly trained on the sampled blank shapes; SAISM, to be noted, should be trained on the latent vectors of the blank shapes and the maximum thinning/thickening values, while the latent vectors should be generated by the trained auto-decoder.

For SAISM, the RBF method is implemented using the Python toolkit scipy.interpolate.Rbf. The Kriging method is implemented using the Python toolkit SMT. All the training parameters retain the default settings. For both RBF and Kriging, two separated models are trained for predicting the maximum thinning and maximum thickening.

For IAISM, all codes were written in Pytorch. The Mask-Res-SE-U-Net was run on an NVIDIA Quadro RTX 5000 GPU and the CPU thread number is 8. The training epoch number was fixed to 2,000. The batch size was set to 4, and the learning rate was set to 0.0004 for the Adam optimiser [74]. The random seed was set to 37. The loss function was defined as:

$$L = \lambda_1 \frac{1}{N}\sum_{i=1}^{N}(y_i - \hat{y}_i)^2 - \lambda_2 \frac{Y \cdot \hat{Y}}{\|Y\|\|\hat{Y}\|}, Y = [y_1, \ldots, y_N], \hat{Y} = [\hat{y}_1, \ldots, \hat{y}_N] \qquad (8)$$

where $L$ is the loss, $N$ is the number of the pixels ($1120 \times 610$ in this research), $Y$ is the ground truth thinning field, $\hat{Y}$ is the predicted thinning field. The first term of $L$ is the mean square error term, and the second term of $L$ is the cosine similarity term. The value of $\lambda_1$ and $\lambda_2$ was empirically set to be 1.0 and 0.2 in this research.

For the auto-decoder, all settings are identical with IAISM, except for the batch size and the $\lambda_1$ value. The batch size is set to be 16, since the training set size of the auto-decoder (1280 samples) is much larger than that of IAISM (256 samples). The value of $\lambda_1$ is set to be 0.01, since the MSE magnitude of the auto-decoder is two or three orders higher than that of IAISM. To be noted, the input layer of latent vectors are learnable parameters of the auto-decoder neural network, as shown in Figure 4 (c).

## 4. Evaluation and discussion

In section 4, the auto-decoder, SAISM (RBF and Kriging), and IAISM (Mask-Res-SE-U-Net) are quantitatively evaluated. To specify, the performance of the auto-decoder is evaluated by the reconstruction accuracy of blank shapes, and also by the interpolation smoothness in the latent space; the performance of SAISM is evaluated by the prediction accuracy of maximum thinning/thickening values, namely the scalar prediction accuracy; the performance of IAISM is evaluated by the scalar prediction accuracy, and also visually evaluated by it's the texture fidelity of the predicted thinning fields, namely the texture prediction accuracy. Based on the trained and evaluated auto-decoder and IAISM, gradient-based optimisation is conducted, according to the pipeline in Figure 3 (a), to search for blank shapes that satisfy the engineering criteria (maximum thinning/thickening lower than 0.15/0.1).

### *4.1 Evaluation of the auto-decoder*

As shown in Figure 4 (c), the trained auto-decoder can be used to either find the latent vector of a given blank shape, or generate the blank shape from a given arbitrary latent vector. Therefore, the trained auto-decoder is evaluated by two criteria:

The first criterion is whether the auto-decoder can find a latent vector that can reconstruct a given blank shape SDF with a high fidelity. This is evaluated on both the training and test sets: for the training set, this evaluation will be directly based on the latent vectors generated after the training process; for each sampled in the test set, a random initial latent vector will be assigned. From the initial vector, Adams optimiser will be applied to, with a learning rate of 0.4 and a training epoch number of 1000, to optimise the latent vector for better reconstructing the blank shape SDF of this test sample. This optimiser setting ensures the latent vectors of the test set samples have similar value ranges with those of the training set samples, as shown in Figure 7 (a). The evaluation of reconstruction accuracy for the test set will be based on these post-optimised latent vectors.

Given that the prediction error of the maximum/minimum SDF is not concerned, two pixel-wise measures are applied to quantify the reconstruction accuracy, namely the maximum absolute pixel-wise error (MAPE) and the averaged absolute pixel-wise error (AAPE):

$$MPAE = \max(|SDF_i^{GT} - SDF_i^{PD}|, i = 1,2,\ldots,N) \quad (9)$$

$$AAPE = \frac{1}{N}\sum_{i=1}^{N}|SDF_i^{GT} - SDF_i^{PD}| \quad (10)$$

where GT refers to ground truth and PD refers to prediction; N refers to the total number of pixels, which is 1120×610 in this research.

For the training set (1280 samples), the averaged MPAE across all samples is 9.79, and the highest MPAE across all samples is 10.69; the averaged AAPE across all samples is 0.247, and the highest AAPE across all samples is 0.295. Given that the SDF values near the image edges can be more than 200, the MPAE and AAPE measures indicate the high reconstruction accuracy of the auto-decoder on the training set, as visualised in Figure 7 (b).

For the test set (64 samples), the averaged MPAE across all samples is 17.23, and the highest MPAE across all samples is 54.37; the averaged AAPE across all samples is 1.14, and the highest AAPE across all samples is 3.40. The reconstruction accuracy of the test set is lower than that of the training set, but still acceptable for reconstruction work, as visualised in Figure 7 (c).

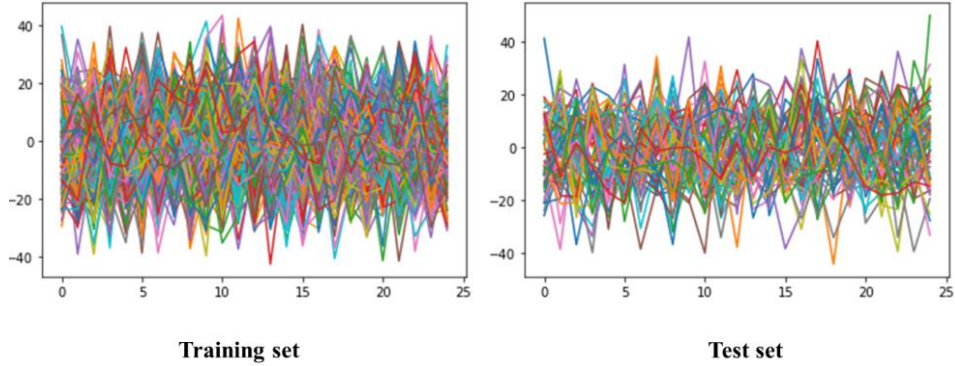

(a)

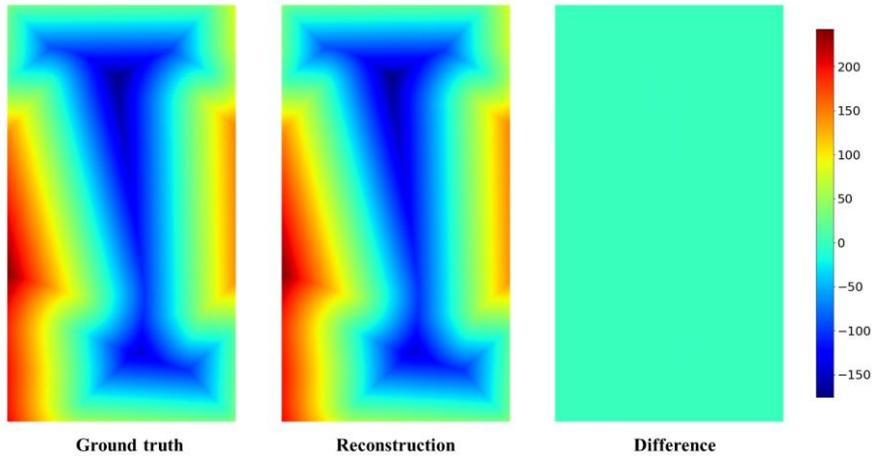

(b)

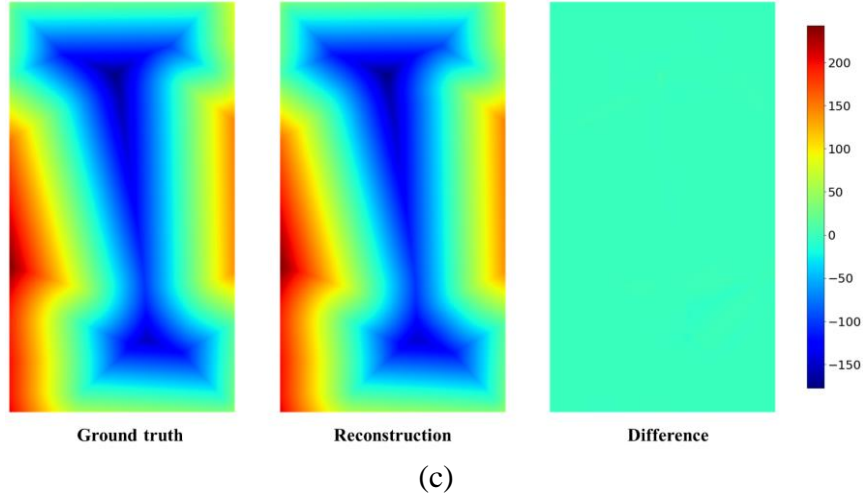

(c)

Figure 7. (a) The value ranges of the 25-D latent vectors in the training/test sets; (b) the reconstruction performance of a sample randomly selected from the training set; (c) the reconstruction performance of a sample randomly selected from the test set.

The second criterion is whether the auto-decoder can smoothly transform a given blank shape to another blank shape by modifying the latent vector. To evaluate the interpolation smoothness, two latent vectors are randomly selected from the training set, namely LV1 and LV2. Eight interpolated latent vectors, namely ILV 1 to 8, are uniformly sampled along the connection line between LV1 and LV2. The ten blank shape SDFs are reconstructed by the auto-decoder. The corresponding blank shapes, which are extracted using the marching cube algorithm [54] from the blank shape SDFs, are visualised in Figure 8. It is observed that the auto-decoder can smoothly modify the blank shape from the left to the right.

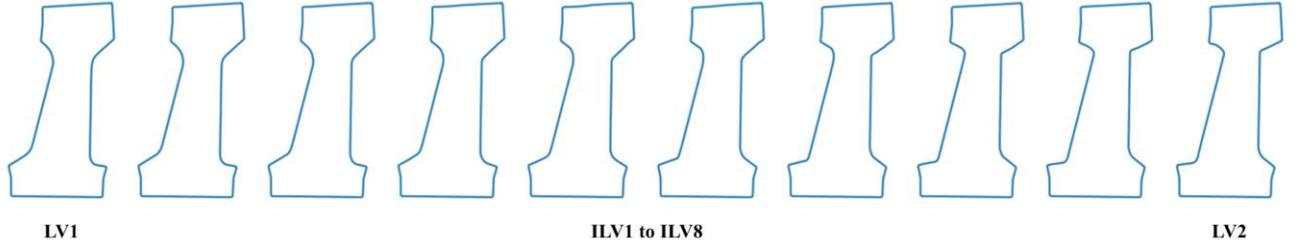

Figure 8. The auto-decoder transforms a given blank shape (leftmost) to another blank shape (rightmost) by modifying the latent vector.

### 4.2 Evaluation of scalar prediction accuracy

The scalar prediction accuracy can be quantified using the relative error of maximum thinning (RMT) and thickening (RMTK):

$$RMT = \frac{abs|Tmax^{GT}-Tmax^{PD}|}{abs|Tmax^{GT}|} \tag{9}$$

$$RMTK = \frac{abs|Tkmax^{GT}-Tkmax^{PD}|}{abs|Tkmax^{GT}|} \tag{10}$$

where $Tmax^{GT}$ denotes the ground truth maximum thinning, $Tmax^{PD}$ denotes the surrogate predicted maximum thinning, $Tkmax^{GT}$ denotes the ground truth maximum thickening, $Tkmax^{PD}$ denotes the surrogate predicted maximum thickening.

For each single test sample, RMT and RMTK are evaluated. A couple of statistics of RMT and RMTK on the test set are calculated to compare the performance of SAISM and IAISM. These statistics include: **ARMT** which is the averaged RMT over the test set, and **ARMTK** which is the averaged RMTK over the test set.

These statistics are collected in Table 3. In total, six SAISM/IAISM models are trained and evaluated, including the RBF predicting the maximum thinning and RBF predicting the maximum thickening, the Kriging predicting the maximum thinning and Kriging predicting the maximum thickening, the SAISM with and without data augmentation. Empirically, a surrogate model with both ARMT and ARMTK under 10% can satisfy the requirements of practical engineering use. Several observations can be obtained from Table 2:

- By comparing MRMT and MRMTK, IAISM has significantly higher accuracy than SAISM (both RBF and Kriging), despite that IAISM does not directly predict the maximum thinning/thickening (but predict the full thinning/thickening field instead).
- By comparing the performance of RBF (model number 1 and 2) and Kriging (model number 3 and 4), it can be seen that the two SAISM methods have similar performance.
- According to the results of model 5 and 6, the performance of IAISM can reach an impressively high level, which satisfies the requirement of industry practice.
- By comparing the performance of models 5 and 6, it is observed that data augmentation slightly improved MRMT, MRMTK in this case study. To be noted, a general conclusion of the impact data augmentation cannot be drawn from these observations, since it requires more benchmark studies and ablation tests.
- By comparing the prediction performance on ARMT and ARMTK, IAISM has significantly better performance on ARMT than ARMTK. This is reasonable because maximum thickening occurs with distorted local distributions like wrinkles, which might increase the hardship of prediction.

Table 2. Performance evaluation of SAISM and IAISM

| Model | RBF | | Kriging | | Mask-Res-SE-U-Net | |
|---|---|---|---|---|---|---|
| Model number | 1 | 2 | 3 | 4 | 5 | 6 |
| Training set size | | | | 256 | | |
| Augmentation | No augmentation available | | | | No | Yes |
| Test set size | | | | 64 | | |
| ARMT | 26.4% | None | 27.9% | None | 4.52% | 4.17% |
| ARMTK | None | 22.2% | None | 26.5% | 8.85% | 8.33% |

*4.3 Evaluation of the texture prediction accuracy*

The thinning fields predicted by IAISM with data augmentation (model 6) are visualized and compared with the ground truth thinning fields in Figure 9, both plotted on the initial blank. In Figure 9 (a), the field predictions of the IAISM model are compared against the ground truth thinning fields calculated in PAM-STAMP: A manufacturability-satisfying sample is selected as test sample 1; A nearly manufacturability-satisfying sample, of which the maximum thinning/thickening values are slightly over the criteria (0.15/0.1), is selected as test sample 2; A severely over-thinned sample is selected as test sample 3; A severely over-thickened sample is selected as test sample 4. In all the four test samples, IAISM reasonably capture the distributions and local textures with a high fidelity, with a better performance in the thinning regions than the thickening regions with wrinkles. In Figure 9 (b), it is observed that the predicted training sample has a better thickening texture fidelity than the predicted test sample; while in the thinning regions, the predicted training and test samples have similar texture fidelity. Figure 9 demonstrates the capability of IAISM to predict high-fidelity physical fields, while also indicates that IAISM generalises better in predicting the thinning regions.

The high-fidelity thinning fields predicted by IAISM provides significant advantages over SAISM. To specify, the users of the surrogate models can hardly evaluate whether the SAISM-predicted maximum thinning/thickening values are accurate enough for design decision making. After all, they are just two scalar values without any extra information. In contrast, IAISM has better interpretability as users can judge whether the prediction is reliable based on the whole physical fields, rather than single values. The users can also evaluate whether the IAISM is well-trained according to their expertise in the underlying physics.

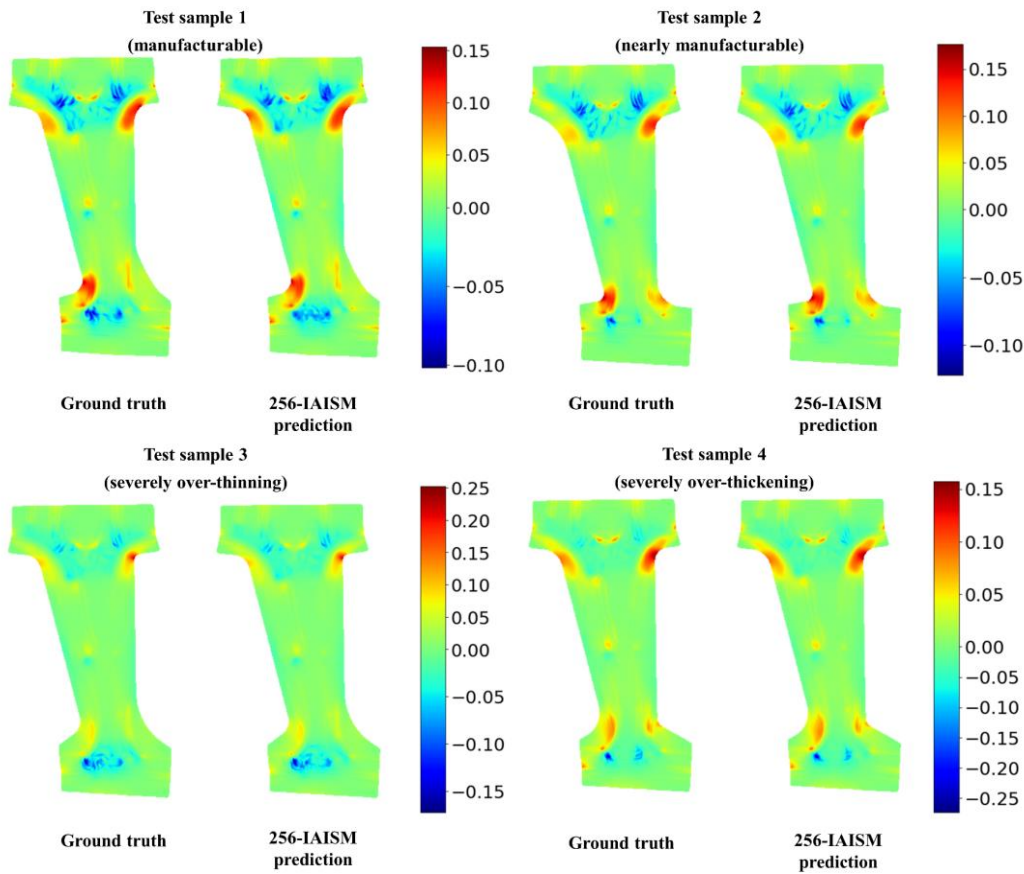

(a)

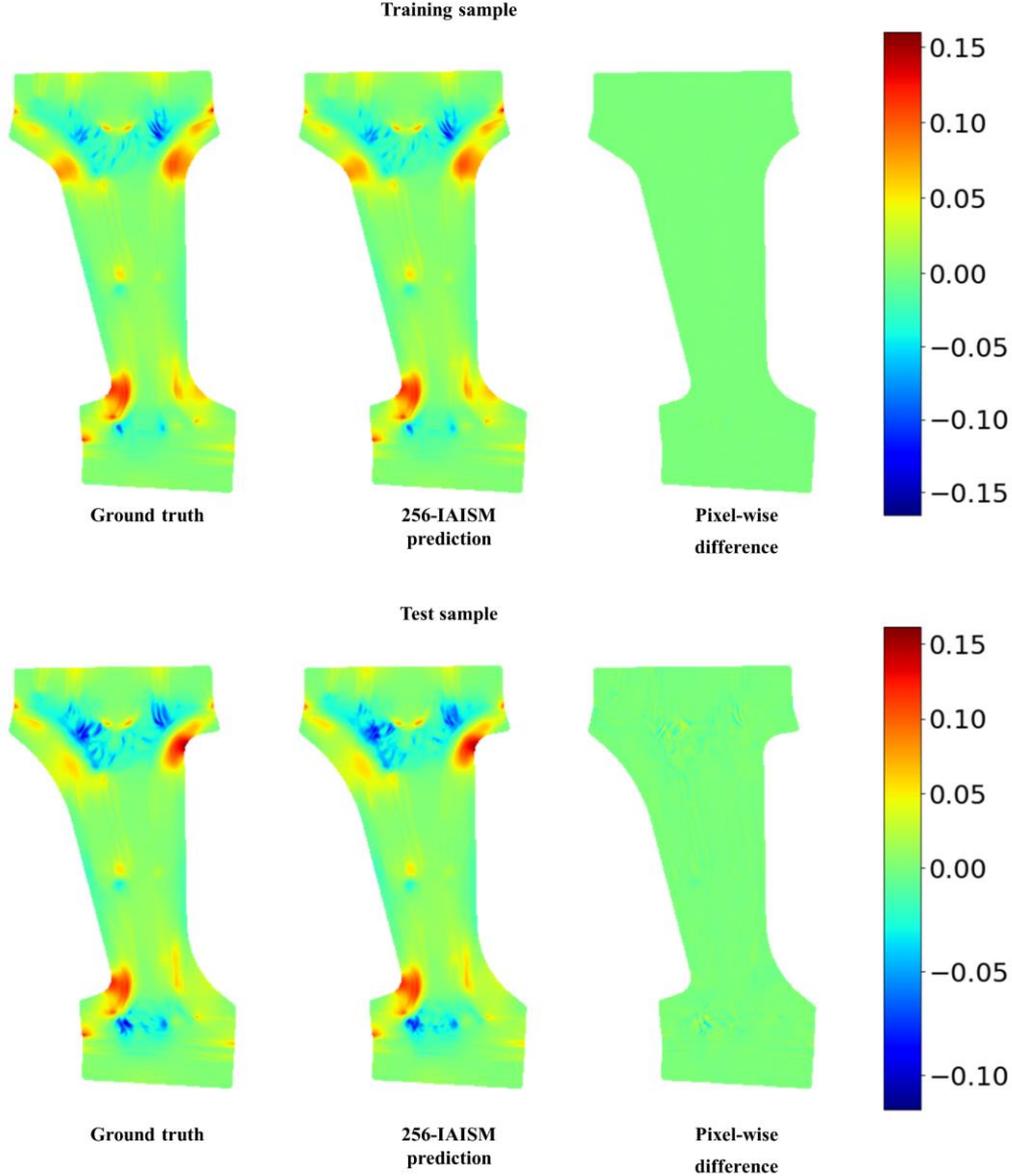

(b)

Figure 9. (a) The IAISM-predicted thinning fields of several test samples; (b) A comparison between the IAISM-predicted thinning fields of a training sample and a test sample.

*4.4 Gradient-based optimisation and validation in PAM-STAMP*

The trained and evaluated IAISM (model 6) is deployed together with the trained auto-decoder to conduct gradient-based optimisation, the pipeline shown in Figure 3 (a). Given a selected initial latent vector, the auto-decoder takes the latent vector and reconstructs the blank shape SDF; IAISM then takes the reconstructed blank shape SDF to generate the thinning field; the maximum thinning/thickening values, which are used to calculate the loss, are extracted from the generated thinning field; the gradient-based optimiser is applied to reduce the loss by modifying the latent vector. This optimisation process iterates until converging to a satisfying latent vector. The loss function is formulated using penalty and regularisation terms as:

$$L = \lambda_1 |maxThickening| + \lambda_2(|maxThinning| - threshold) + \lambda_3 line \qquad (11)$$

where $maxThickening$ and $maxThinning$ are extracted from the IAISM-generated field; $\lambda_1, \lambda_2, \lambda_3$ are tuneable coefficients. Since predicting the maximum thickening is more difficult, as discussed in section 4.3, the maximum thickening term (first term) is set to be the objective term, and the maximum thinning term (second term) is set to be the penalty term with a $threshold$ coefficient. $line$ is a regularisation term to keep the edge in red-circled region 1 in Figure 5 (a) as a straight line with a fixed slope during deformation. $line$ is calculated as the averaged gradient norm difference of the local SDF, as shown in Figure 10:

$$line = \frac{1}{M}\sum_{i=1}^{M}((G_x^i(SDF) - 0.0603)^2 + (G_y^i(SDF) - 0.9982)^2)$$

where $G_x^i$ is the X-direction gradient value of the local i-th pixel, and 0.0603 is the reference value; $G_y^i$ is the Y-direction gradient value of the local i-th pixel, and 0.9982 is the reference value; M is the total pixel number in the local area (boxed in Figure 10).

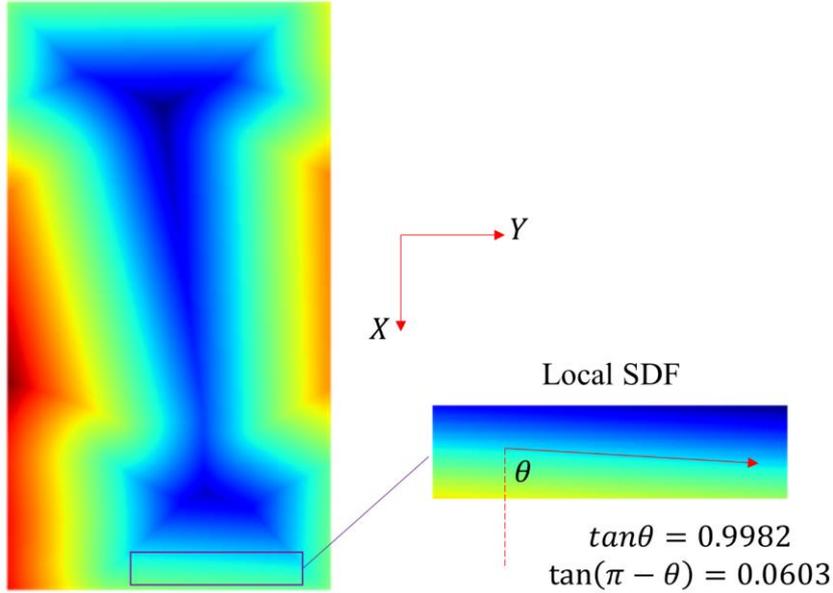

Figure 10. The definition of the regularisation term in Eq. (11).

Empirically, $\lambda_1$ is assigned to be 0.1, $\lambda_2$ is set to be 0.35, $\lambda_3$ is assigned to be 1.5, and $threshold$ is assigned to be 0.13. An empirical rule of the starting latent vector is to select a training sample with a low maximum thickening value, since predicting and optimising the maximum thickening is expected to be more difficult, as discussed in section 4.3. In our research, two optimisation case studies are conducted. The Adam optimiser is applied with a learning rate of 2.0 and an optimisation epoch number of 2000. After the optimisation, the auto-decoder will reconstruct the blank shape SDF and extract the blank shape using the marching cube algorithm.

For each case study, the effectiveness of optimisation is demonstrated in Figure 11 (a). It is observed that the optimiser intelligently modifies the local blank shape features to reduce both the IAISM-predicted maximum thinning and thickening values. To further validate the effectiveness of the optimisation, the optimised blank shape will be extracted and evaluated in PAM-STAMP. For both cases, the ground truth thinning fields of both cases satisfy the manufacturability criteria (0.15/-0.10), as shown in Figure 11 (b). To be noted, no sample in the training set satisfy manufacturability criteria, which indicates the difficulty of this blank shape optimisation problem. Despite of the difficulty, the Adam optimiser found the satisfying blank shapes within 10 min in both cases. This demonstrates the impressive capability of the optimisation pipeline in our research.

**Case 1**

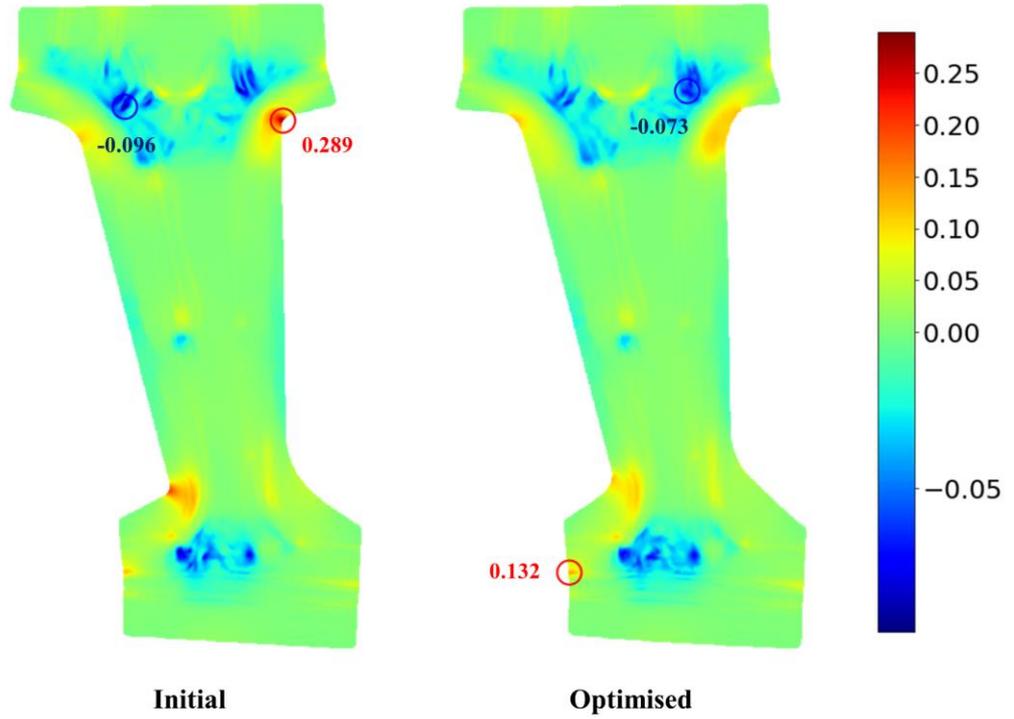

Initial                               Optimised

**Case 2**

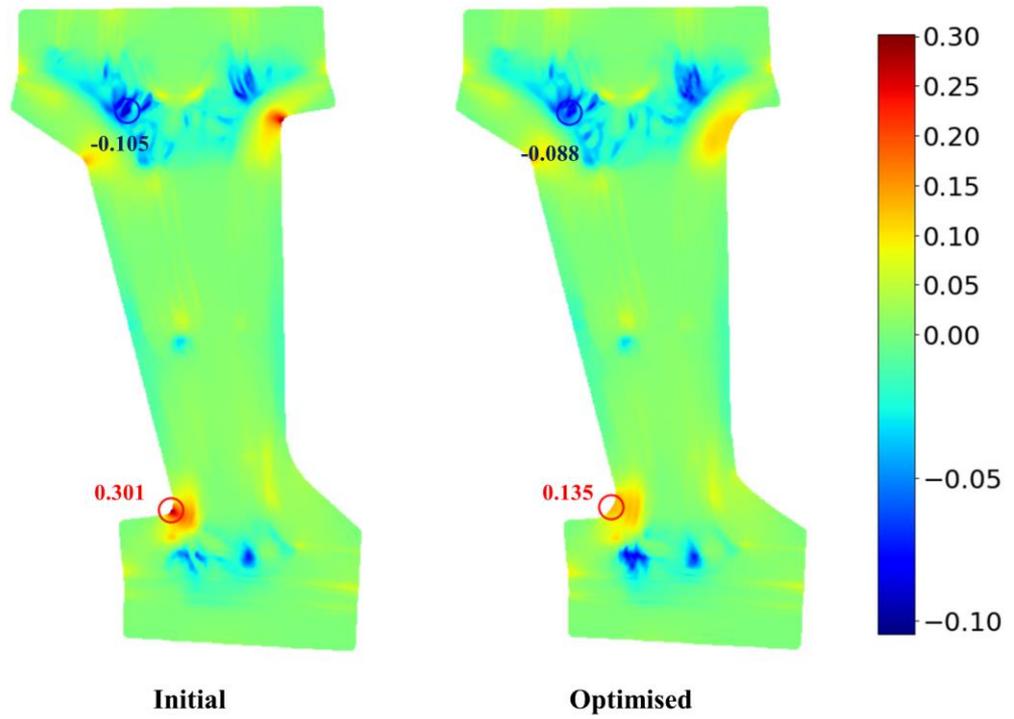

Initial                               Optimised

(a)

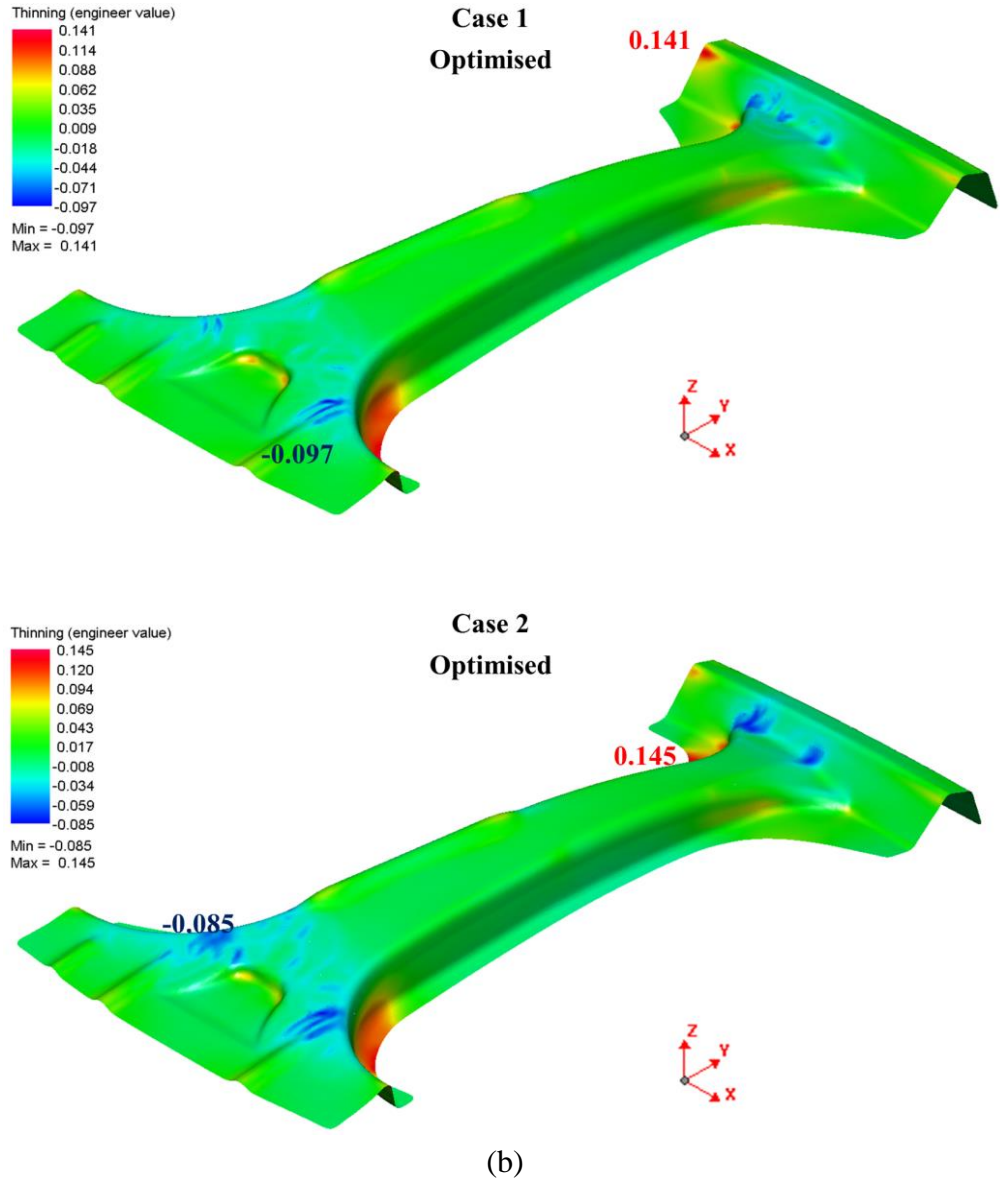

(b)

Figure 11. (a) The comparison between the initial thinning field and optimised thinning field for each case. To be noted, all the circled values refer to the IAISM-predicted values, rather than the corresponding ground truth in PAM-STAMP simulation. (b) the PAM-STAMP simulation results of the optimised blank shapes in the two cases.

## 5. Conclusions

In this research, the blank shape of a hot-stamped B-pillar for an automotive application was parameterised in sixteen ways. Under these parameterisation, two training sets (1280 samples for the auto-decoder and 256 samples for SAISM/IAISM) and one test set (64 samples for the auto-decoder, SAISM/IAISM) were generated using a specialized data sampling strategy. The auto-decoder was trained and evaluated based on the criteria of reconstruction accuracy and interpolation smoothness. The trained auto-decoder was able to represent an arbitrary blank shape SDF as a latent vector, or reconstruct the blank shape SDF from an arbitrary latent vector; SAISM and IAISM were trained and evaluated to investigate the advantages of IAISM. RBF and Kriging were selected as the state-of-the-art and representative SAISM methods, and Mask-Res-SE-U-Net was developed as the IAISM in this research. SAISM and IAISM, in total six models (RBF for maximum thinning/thickening, Kriging for maximum thinning/thickening, Mask-Res-SE-U-Net, Mask-Res-SE-U-Net with data augmentation), were trained and evaluated, respectively. The performances of SAISM and IAISM are evaluated and compared based on the accuracy of predicted maximum thinning/thickening values in the test set. According to the evaluation results, IAISM significantly outperformed SAISM in predicting maximum thinning/thickening values, even though these values were not explicitly formulated in the loss function of IAISM. The averaged relative errors of IAISM (ARMT lower than 5% and ARMTK lower than 10%) were significantly lower than those of SAISM (ARMT higher than 25% and ARMTK higher than 20%). Moreover, IAISM predicted full thinning fields with high-fidelity texture, which grants IAISM better interpretability.

After training and evaluation, the auto-decoder, IAISM (with data augmentation), and Adam optimiser were integrated to conduct gradient-based optimisation, in order to improve manufacturability by modifying the latent vector. In two case studies, the optimiser successfully found a blank shape that satisfied the manufacturability criteria (0.15/-0.10) within 10 min, which was validated in PAM-STAMP.

The above findings indicate the significant and quantitative advantages of IAISM over SAISM, which can be partly attributed to the SDF-field shape representations and image-based simulation representations. With proper data representations, deep learning based surrogate models can achieve high accuracy and texture fidelity trained on small datasets, which indicates the special attributes of engineering AI. Instead of following the stereotype of big data, new theories and methodologies should be proposed and investigated for engineering AI task. This research validates the effectiveness of gradient-based optimisation equipped with high-fidelity IAISM and shape generator (auto-decoder in this case). This research provides a high-fidelity, widely-generalizable, and industry-applicable surrogate optimisation solution.

**Appendix A**

- For region 1, the edge can be moved along its normal direction, while keeping the edge straight. As shown in the region 1 block in Figure 5 (b), this parameterisation is defined using a parameter P0, which defines the distance of the upper edge to a fixed dashed subline.
- For regions 2 and 3, the local transition curve can either be an arc or a spline with two control points. If parameterised as an arc, P1 and P8 define the angles of the upper small arcs, respectively for region 2 and region 3. (Note: in the following context, the former parameter always depicts region 2 and the latter depicts region 3). P2 and P9 define the radii of the main arcs. If parameterised as a spline with two control points, each spline contains two control points and two end points which are attached to the upper and lower solid straight lines. P3 and P10 define the angles of the upper small arcs. P4 and P11 define the lengths of the upper straight lines. P5 and P12 define the lengths of the upper sublines, which are respectively on the extension of the solid straight lines defined by P4 and P11. The sublines points defined by P5 and P12 are upper control points. Two auxiliary sublines that are normal to the lower solid straight lines (the green-circled edges in Figure 5 (a)), respectively in region 2 and 3, are added to define the lower spline end points. P6 and P13 define the distances from the upper spline end points to the two auxiliary sublines, which then defines the lower spline end points. P7 and P14 define the lengths of the lower sublines, which are respectively on the extension of the lower solid straight lines. The end points of the sublines defined by P6 and P13 are lower control points.
- For regions 4 and 5, the local transition curve can either be an arc or a spline with three control points. If parameterised as an arc, P15 and P24 define the angles of the lower small arcs, respectively for region 4 and region 5. (Note: in the following context, the former parameter always depicts region 4 and the latter depicts region 5). P16 and P25 define the radii of the main arcs. If parameterised as a spline with three control points, each spline contains three control points and two end points, which are attached to the upper and lower solid straight lines. P17 and P26 define the angles of the lower small arcs. P18 and P27 define the lengths of the lower straight lines. P19 and P28 define the lengths of the lower sublines, which are respectively on the extension of the solid straight lines defined by P18 and P27. The end points of the sublines defined by P19 and P28 are lower control points. Two auxiliary sublines that are normal to the upper solid straight lines (the green-circled edges in Figure 5 (a)), respectively in region 4 and 5, are added to define the upper end points of the splines. P21 and P29 define the distances from the centres of the lower small arcs to the two auxiliary sublines, which then defines the upper spline control points. P20 and P30 define the lengths of the upper sublines, which are respectively on the extension of the upper solid straight lines. The end points of the sublines defined by P20 and P30 are upper control points. The middle spline control points in regions 4 and 5 are defined using their relative positions to the corresponding centres of the lower small arcs. To specify, the middle spline control point in region 4 is defined using P22 and P23, and the middle spline control point in region 5 is defined using P31 and P32.

**Appendix B**

- P0 can be a RI parameter. The range of P0 is set to be [10mm, 70mm].
- P1, P3, P8, P10, P15, P17, P24, P26, as the angles of the small arcs, can be RI parameters.. The ranges for P1, P3, P8, P10 are set to be [50deg, 90deg]. The ranges for P15, P17, P24, P26 are set to be [60deg, 100deg].
- P2, P9, P16, P25, as the radii of the main arcs, are RD parameters since straight lines between the small arcs and the main arcs should have positive lengths. In this research, the lower limits of these four radii parameters are 15 mm, while the upper limits should guarantee the minimum lengths of the in-between lines to be 15 mm.
- P4, P11, P18, P27, as the lengths of the four in-between straight lines, are RD parameters. For each in-between straight line, the lower limit is set to be 15mm to avoid a too small length. The upper limit should avoid the intersection between the straight line, which link the small arc and the spline, and the extension of the straight edge (green-circled in Figure 5 (a)). In practice, a 15mm allowance is applied to avoid interaction.
- P5, P7, P12, P14, P19, P20, P28, P30, as the lengths of the spline sublines, are RD parameters. For each spline subline, the lower limit of length is set to be 5 mm. The upper limit should prevent the upper and lower spline control points from crossing the interaction point between the extension of the straight line, which links the small arc and the spline, and the extension of the straight edge (green-circled in Figure 5 (a)). In practice, a 2mm allowance is applied to prevent the spline control points from crossing the intersection point.
- P21 and P29, as the distance from the small arc centres to the auxiliary sublines, are RD parameters. For each parameter, the lower limit should guarantee a minimum of length of 15 mm from the auxiliary subline to the upper spline control point. Since the straight edges (green-circled in Figure 5 (a)) are long enough, there are no restrictions on the upper limit. Empirically, the upper limit is set to allow a maximum length of 100 mm from the auxiliary subline to the upper spline control point.
- P22, P23, P31, P32, as the coordinates of the middle spline control points, are RD parameters. For each middle spline control point, the two parameters are empirically set to restrict the middle points within a semi-circle, which is towards the blank interior and its diameter is from the upper to the lower spline control point.